\pdfoutput=1

\documentclass[11pt]{article}

\usepackage[table]{xcolor}

\usepackage[final]{acl}

\usepackage{times}
\usepackage{latexsym}

\usepackage[T1]{fontenc}

\usepackage[utf8]{inputenc}

\usepackage{microtype}

\usepackage{inconsolata}

\usepackage{graphicx}

\usepackage{multirow}
\usepackage{array}

\usepackage{quoting}
\usepackage{caption}

\usepackage{amsmath}
\usepackage{tcolorbox}
\usepackage{enumitem}

\usepackage{newunicodechar}

\usepackage{CJKutf8}
\newcommand{\Chinese}[1]{{\begin{CJK*}{UTF8}{gbsn}#1\end{CJK*}}}
\newcommand{\Japanese}[1]{{\begin{CJK*}{UTF8}{min}#1\end{CJK*}}}

\title{Multilingual and Continuous Backchannel Prediction:\\A Cross-lingual Study}

\author{
 \textbf{Koji Inoue},
 \textbf{Mikey Elmers},
 \textbf{Yahui Fu},
 \textbf{Zi Haur Pang},
 \textbf{Taiga Mori},\\
 \textbf{Divesh Lala},
 \textbf{Keiko Ochi},
 \textbf{Tatsuya Kawahara}
\\
\\
 Graduate School of Informatics, Kyoto University, Japan, \\
\\
 \small{
   \textbf{Correspondence:} \href{mailto:inoue@sap.ist.i.kyoto-u.ac.jp}{inoue@sap.ist.i.kyoto-u.ac.jp}
 }
}

\begin{document}

\maketitle

\begin{abstract}
We present a multilingual, continuous backchannel prediction model for Japanese, English, and Chinese, and use it to investigate cross-linguistic timing behavior.
The model is Transformer-based and operates at the frame level, jointly trained with auxiliary tasks on approximately 300 hours of dyadic conversations.
Across all three languages, the multilingual model matches or surpasses monolingual baselines, indicating that it learns both language-universal cues and language-specific timing patterns.
Zero-shot transfer with two-language training remains limited, underscoring substantive cross-lingual differences.
Perturbation analyses reveal distinct cue usage: Japanese relies more on short-term linguistic information, whereas English and Chinese are more sensitive to silence duration and prosodic variation; multilingual training encourages shared yet adaptable representations and reduces overreliance on pitch in Chinese.
A context-length study further shows that Japanese is relatively robust to shorter contexts, while Chinese benefits markedly from longer contexts.
Finally, we integrate the trained model into a real-time processing software, demonstrating CPU-only inference.
Together, these findings provide a unified model and empirical evidence for how backchannel timing differs across languages, informing the design of more natural, culturally-aware spoken dialogue systems.
\end{abstract}

\renewcommand{\thefootnote}{\fnsymbol{footnote}}
\footnote[0]{This paper has been accepted for presentation at International Workshop on Spoken Dialogue Systems Technology 2026 (IWSDS 2026) and represents the author’s version of the work.}
\renewcommand{\thefootnote}{\arabic{footnote}}

\section{Introduction}

Smooth human conversation is supported by brief listener responses such as ``uh-huh'' and ``oh,'' known as \textit{backchannels}, produced at appropriate moments~\cite{schegloff1982discourse,clark1996using,clancy1996conversational}.
Backchannels serve not only to regulate turn-taking but also to signal interest and understanding, and are thus essential for spoken dialogue systems that aim to interact in a human-like manner~\cite{schroder2011building,devault2014simsensei,inoue2016erica}.
Their importance is also recognized in emerging full-duplex spoken dialogue systems, for which modeling and evaluation foundations are being established~\cite{moshi,lin2025fullduplexbenchv15evaluatingoverlap}.

Automatic backchannel generation involves predicting three factors: timing, form, and prosody.
Among these, \textit{timing}—when to produce a backchannel—is fundamental.
Prior work has explored utterance-level and frame-level (continuous) prediction~\cite{jang2021bpm_mt,ruede2017yeah}.
Because humans often insert backchannels before a speaker’s utterance is complete, continuous frame-level models are preferable for reproducing human-like behavior.
However, continuous models face challenges such as label imbalance.
Recent approaches improve performance via multi-task learning with related tasks such as turn-taking prediction~\cite{hara2018multi,choi2024etri,inoue2025naacl}.

Most backchannel studies to date have targeted a single language (often Japanese or English), and cross-linguistic analyses remain limited.
Backchannel frequency and timing vary by language and culture; for example, in Japanese, backchannels often occur during the speaker’s ongoing utterance, whereas in Chinese they more frequently appear after utterance completion~\cite{clancy1996conversational}.
Quantifying these differences and modeling both universal and language-specific features are crucial steps toward dialogue technologies that are robust across diverse linguistic cultures.

\begin{table*}[t]
\centering
\caption{Statistics of backchannel data by language}
\label{tab:bc_stats}
\setlength{\tabcolsep}{5mm}
\begin{tabular}{lrrr}
\hline
& \multicolumn{1}{c}{\textbf{Japanese}} & \multicolumn{1}{c}{\textbf{English}} & \multicolumn{1}{c}{\textbf{Chinese}} \\
\hline
\# Dialogues & 299 & 300 & 298 \\
Total dialogue time & 108:13:34 & 119:56:12 & 108:05:12 \\
Total used dialogue time & 49:13:39 & 27:20:31 & 25:04:53 \\
\cline{2-4}
\# Backchannel utterances & 58800 (34.4\%) & 24612 (28.4\%) & 21182 (27.5\%) \\
\# Non-backchannel utterances & 112177 & 62158 & 55955 \\
\cline{2-4}
Total BC time [s] & 29253.73 (16.5\%) & 11006.73 (11.2\%) & 7695.41 (8.5\%) \\
Total non-BC time [s] & 147965.49 & 87424.67 & 82598.55 \\
\hline
\end{tabular}
\end{table*}

To this end, we conduct a comparative analysis of backchannel timing in Japanese, English, and Chinese.
We first compile a large-scale, 300-hour multilingual conversational dataset.
We then build a Transformer-based multilingual backchannel prediction model that continuously outputs frame-level probabilities.
The model is designed to learn features that are shared across languages while also capturing language-specific patterns.
Finally, we compare monolingual and multilingual settings and analyze which input aspects are important for predicting backchannels across languages, highlighting commonalities and differences.

\section{Dataset} \label{sec:dataset}

We analyze first-encounter dyadic conversations recorded over an online conferencing tool (Zoom).
The total recording time is nearly 300 hours: about 100 hours each for Japanese, English, and Chinese.
Utterances were segmented manually into Inter-Pausal Units (IPUs) using a 200~ms silence threshold.
We then applied automatic speech recognition (ASR) to each segment to obtain transcripts.
Whisper was used for ASR: {\it kotoba-tech/kotoba-whisper-v2.2}\footnote{\url{https://huggingface.co/kotoba-tech/kotoba-whisper-v2.2}} for Japanese, and {\it large-v3}\footnote{\url{https://huggingface.co/openai/whisper-large-v3}} for English and Chinese.

Using a manually curated surface-form list of backchannels, we identified backchannel utterances from the ASR outputs.
Following prior works~\cite{choi2024etri,inoue2025naacl}, our target forms comprise interjections from the \textit{continuer} class (e.g., ``\Japanese{うん},'' ``yeah,'' and ``\Chinese{嗯}'') and the \textit{assessment} class (e.g., ``\Japanese{へー},'' ``wow,'' and ``\Chinese{哦}'').
For each language, the list was verified by native-speaking authors and consolidated to account for variants and dialectal forms.
Consecutive backchannels, such as ``yeah yeah,'' were merged into a single instance.
If the preceding utterance was produced by the same person who produced the candidate backchannel, we filtered it out (i.e., it was not treated as a listener backchannel).

Since VAP training operates on 20-second segments, we split the dialogues accordingly.
In each segment, the participant who produced more backchannels was labeled the \textit{listener}, and the other the \textit{speaker}.
We then predicted backchannels for the listener.
Segments with no backchannels were excluded.

Table~\ref{tab:bc_stats} summarizes the annotations.
Japanese shows the highest backchannel rate: approximately 34.4\% of utterances and 16.5\% of total time.
English and Chinese exhibit lower rates (28.4\% / 11.2\% and 27.5\% / 8.5\%, respectively), suggesting cross-linguistic differences in backchannel behavior.
The higher frequency in Japanese aligns with prior reports~\cite{maynard1986back,clancy1996conversational,miller2011verbal}.

\begin{table}[t]
\centering
\caption{Share of backchannels occurring during vs.\ after the preceding utterance}
\label{tab:within_after_transposed}
\begin{tabular}{lrr}
\hline
& \multicolumn{1}{c}{\textbf{During utt.}} & \multicolumn{1}{c}{\textbf{After utt.}} \\
\hline
Japanese & 40804 (69.4\%) & 17996 (30.6\%) \\
English & 14981 (60.9\%) & 9631 (39.1\%) \\
Chinese & 10038 (47.4\%) & 11144 (52.6\%) \\
\hline
\end{tabular}
\end{table}

\begin{figure}[t]
\centering
\includegraphics[width=\linewidth]{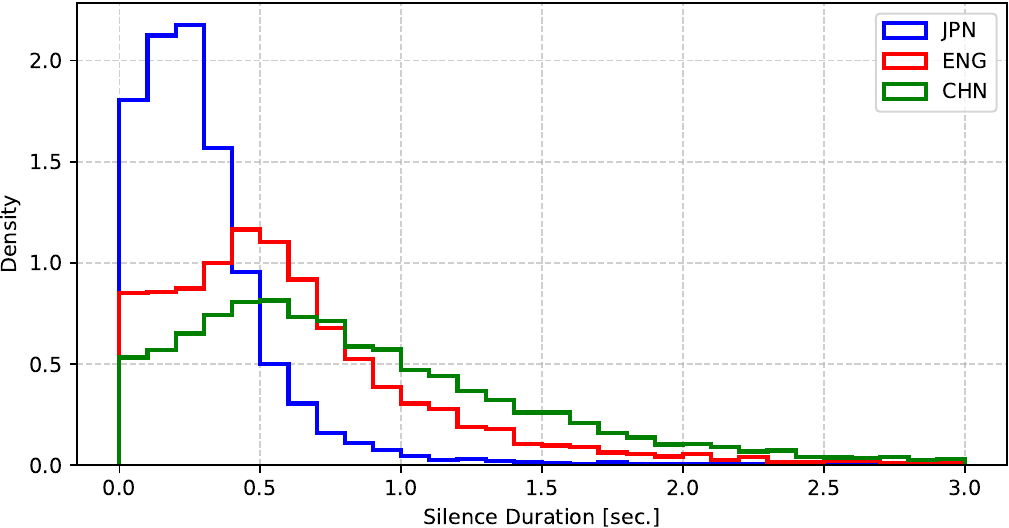}
\caption{Probability density histograms of the time lag between the end of the preceding utterance and the onset of the backchannel}
\label{fig:silence_dist}
\end{figure}

We further analyzed whether backchannels overlap with the speaker’s ongoing utterance or occur after a silence following utterance completion.
As shown in Table~\ref{tab:within_after_transposed}, 69.4\% of Japanese backchannels occur during the speaker’s utterance, which is higher than in Chinese (47.4\%).
This reflects a conversational tendency in Japanese to insert supportive responses mid-utterance, with such insertions less likely to be perceived as interruptions.
By contrast, in Chinese, 52.6\% of backchannels occur after the speaker’s utterance, indicating a preference to respond following clear completion; this points to more explicit turn boundaries and a sharper separation of speaker and listener roles.

Figure~\ref{fig:silence_dist} plots the probability density of post-utterance silence (the time from the end of the preceding utterance to the start of a backchannel).
The peak for Japanese is around 0.2-0.3~s, while English and Chinese both peak near 0.5~s.
Thus, Japanese backchannels tend to occur after shorter silences.
Comparing English and Chinese, the Chinese distribution has a heavier tail with a longer mean silence (about 0.9~s vs.\ about 0.6~s for English), suggesting that Chinese backchannels are more sensitive to the duration of silence.

\section{Backchannel Prediction Model} \label{sec:model}

\begin{figure}[t]
\includegraphics[width=\linewidth]{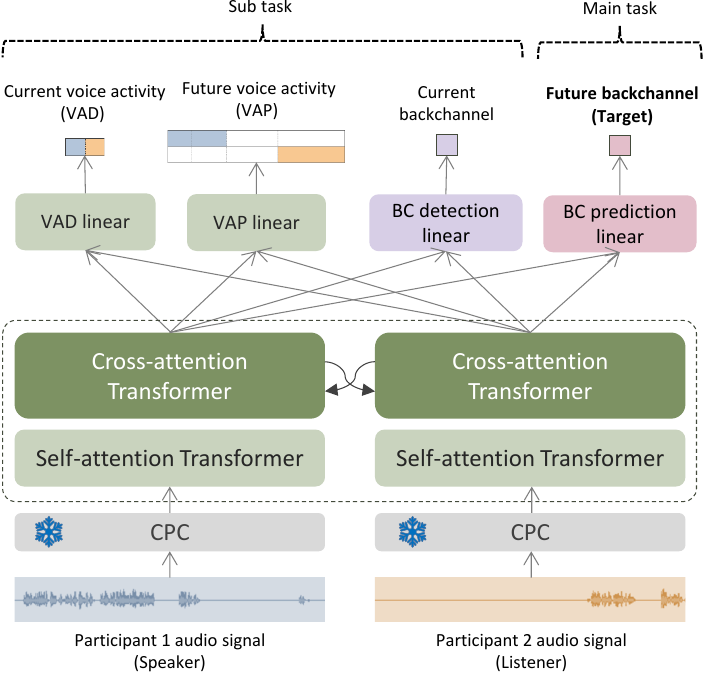}
\caption{Architecture of the backchannel prediction model}
\label{fig:vap}
\end{figure}

We build upon the Voice Activity Projection (VAP) model~\cite{erik2022vap,inoue2025naacl}, which supports continuous prediction (Figure~\ref{fig:vap}).
The inputs are the separated waveforms of two interlocutors—one for the \textit{speaker} and the other for the \textit{listener}.
Each input is encoded by a Contrastive Predictive Coding (CPC) encoder into a feature sequence.
We use a CPC model pre-trained on the Libri-light dataset (about 60k hours)~\cite{riviere2020unsupervised} and keep its parameters frozen.

Encoded features are first processed by separate Transformers and then fused via a Cross-Attention Transformer to capture inter-speaker interactions.
The resulting representation is fed to four linear heads (top of Figure~\ref{fig:vap}), each corresponding to a subtask.
Following \cite{erik2022vap} and \cite{inoue2025naacl}, we design the tasks, which are helpful for stabilizing learning under label imbalance (backchannels can be sparse in some languages), as follows:

\begin{figure}[t]
\includegraphics[width=\linewidth]{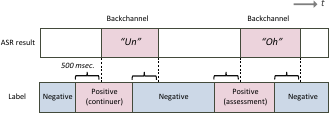}
\caption{Definition of the target label window}
\label{fig:def:bc}
\end{figure}

\begin{figure*}[t]
\centering
(a) Japanese
\includegraphics[width=\linewidth]{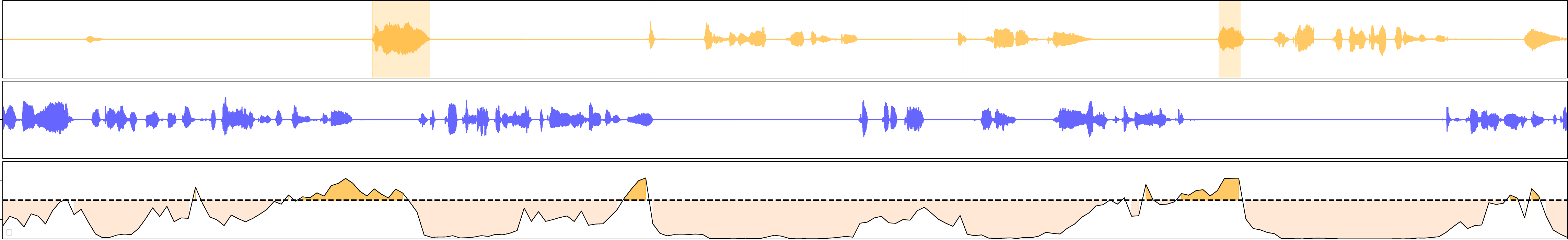}
\vspace{-1mm}\\
(b) English
\includegraphics[width=\linewidth]{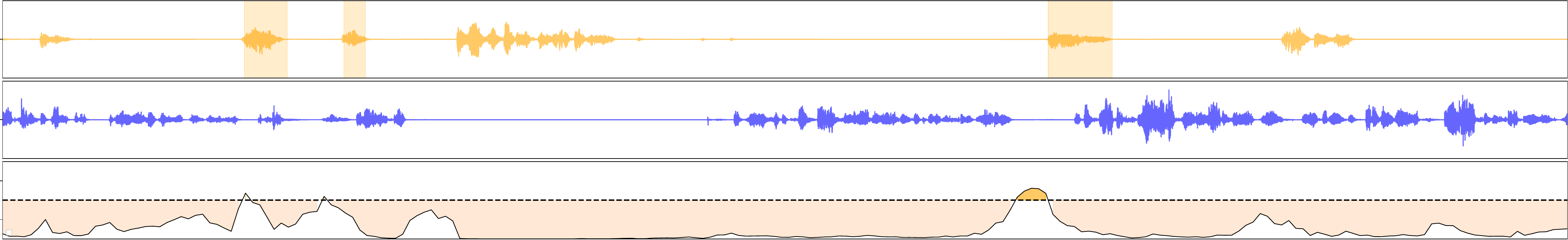}
\vspace{-1mm}\\
(c) Chinese
\includegraphics[width=\linewidth]{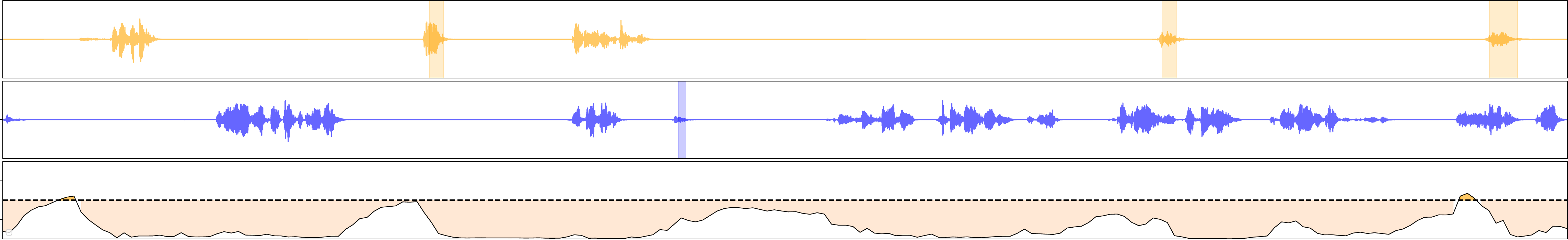}
\caption{Examples of multilingual model behavior on test data for each language (top to bottom in each panel: listener waveform, speaker waveform, listener backchannel prediction probability; highlighted regions indicate backchannel intervals).}
\label{fig:case_study}
\end{figure*}

\begin{itemize}[topsep=2mm]
    \setlength{\itemsep}{0.2em}
    \item \textbf{Voice Activity Detection (VAD)} estimates the probability of speaking vs.\ non-speaking for each interlocutor at the current frame.
    This is the subtask in the original VAP model.

    \item \textbf{Voice Activity Projection (VAP)} predicts the joint speaking states of both interlocutors over the next 2~s, as a proxy for turn-taking prediction.
    We discretize into four bins: 0-200~ms, 200-600~ms, 600-1200~ms, and 1200-2000~ms, and represent the joint state (speak/non-speak for each person) in each bin, yielding a 256-class output.

    \item \textbf{Backchannel Detection (BD)} estimates whether the listener is currently producing a backchannel.
    This task is expected to supplement backchannel prediction by explicitly identifying backchannel instances.

    \item \textbf{Backchannel Prediction (BP)} estimates whether the listener will produce a backchannel 0.5~s in the future.
    This is our main task.
    During training, we shift annotated backchannel onsets by 0.5~s (Figure~\ref{fig:def:bc}) to create supervision targets.
\end{itemize}

The overall loss is
\begin{align}
L = \alpha_1 L_{\text{VAD}} + \alpha_2 L_{\text{VAP}} + \alpha_3 L_{\text{BD}} + \alpha_4 L_{\text{BP}} ~,
\end{align}
where $L_{\text{VAD}}$, $L_{\text{VAP}}$, $L_{\text{BD}}$, and $L_{\text{BP}}$ are the losses for VAD, VAP, backchannel detection, and backchannel prediction, respectively.
We set $\alpha_1 = \alpha_2 = 1.0$ and $\alpha_3 = \alpha_4 = 5.0$ to emphasize the backchannel-related tasks, following \cite{inoue2025naacl}.

\begin{figure*}[t]
\centering
(a) Japanese
\includegraphics[width=\linewidth]{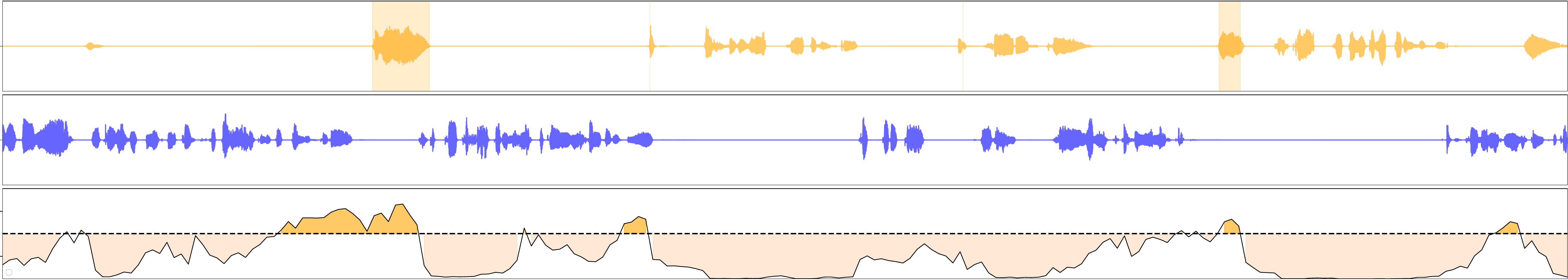}
\vspace{-1mm}\\
(b) English
\includegraphics[width=\linewidth]{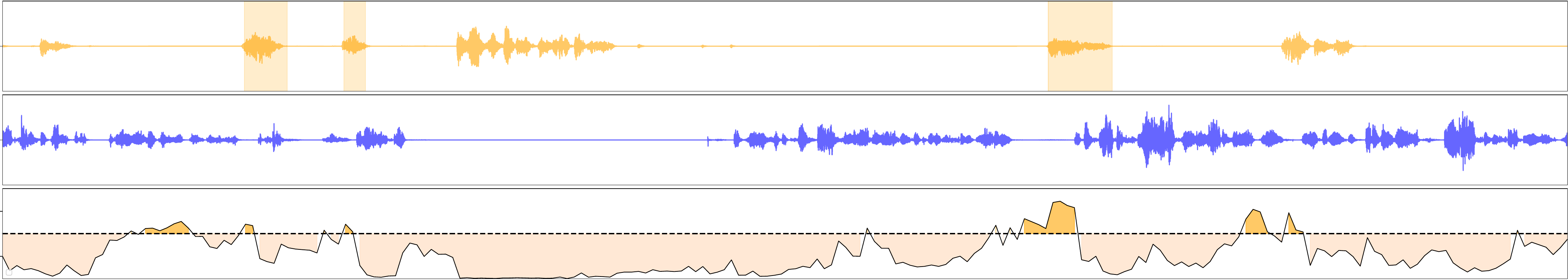}
\vspace{-1mm}\\
(c) Chinese
\includegraphics[width=\linewidth]{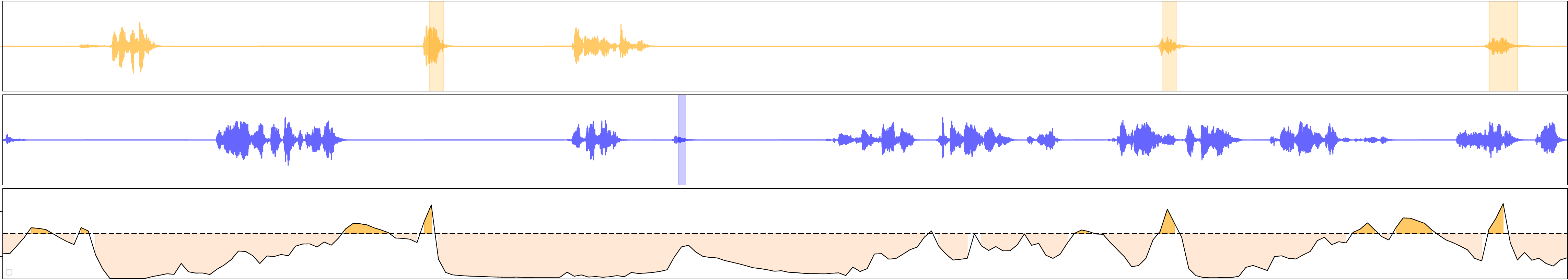}
\caption{Examples of Japanese monolingual model behavior on test data for each language (top to bottom in each panel: listener waveform, speaker waveform, listener backchannel prediction probability; highlighted regions indicate backchannel intervals).}
\label{fig:case_study_jp}
\end{figure*}

\section{Experiments}

We evaluate the proposed multilingual backchannel prediction model and analyze cross-linguistic differences.

\begin{table}[t]
\centering
\caption{Cross-lingual backchannel prediction performance (F1 score [\%])}
\label{tab:f1_scores}
\begin{tabular}{lrrr}
\hline
\multicolumn{1}{c}{\multirow{2}{*}{Training}} & \multicolumn{3}{c}{Test} \\
\cline{2-4}
& \multicolumn{1}{c}{Japanese} & \multicolumn{1}{c}{English} & \multicolumn{1}{c}{Chinese} \\
\hline
Japanese & 33.27 & 15.41 & 10.32 \\
English & 7.92 & 22.85 & 19.52 \\
Chinese & 10.32 & 19.52 & 21.37 \\
Multilingual & \textbf{33.69} & \textbf{23.96} & \textbf{22.65} \\
\hline
\end{tabular}
\end{table}

\begin{table}[t]
\centering
\caption{Zero-shot performance of two-language models}
\label{tab:two_lang}
\begin{tabular}{llc}
\hline
\multicolumn{1}{c}{Training} & \multicolumn{1}{c}{Test} & \multicolumn{1}{c}{F1 score [\%]} \\
\hline
English-Chinese & Japanese & \phantom{0}8.02 \\
Chinese-Japanese & English & 12.33 \\
Japanese-English & Chinese & 17.02 \\
\hline
\end{tabular}
\end{table}

\begin{table*}[t]
\centering
\caption{Ablation result of (matched) \textbf{monolingual} models (F1 score [\%] and drop)}
\label{tab:ablation_mono}
\begin{tabular}{lcrrcrrcrr}
\hline
\multicolumn{1}{c}{Ablation} & & \multicolumn{2}{c}{Japanese} & & \multicolumn{2}{c}{English} & & \multicolumn{2}{c}{Chinese} \\
\hline
Original & & 33.27 & & & 22.85 & & & 21.37 & \\
w/o $L_{\text{BD}}$  & & 33.14 & ($-$0.13) & & 22.91 & ($+$0.06) & & 22.67 & ($+$1.30) \\
w/o $L_{\text{VAP}}$ & & 33.09 & ($-$0.18) & & 22.04 & ($-$0.81) & & 21.22 & ($-$0.15) \\
w/o $L_{\text{VAD}}$ & & 33.81 & ($+$0.54) & & 21.75 & ($-$1.10) & & 22.02 & ($+$0.65) \\
\hline
\end{tabular}

\vspace{15pt}

\caption{Ablation result of \textbf{multilingual} model (F1 score [\%] and drop)}
\label{tab:ablation_multi}
\begin{tabular}{lcrrcrrcrr}
\hline
\multicolumn{1}{c}{Ablation} & & \multicolumn{2}{c}{Japanese} & & \multicolumn{2}{c}{English} & & \multicolumn{2}{c}{Chinese} \\
\hline
Original & & 33.69 & & & 23.96 & & & 22.65 & \\
w/o $L_{\text{BD}}$  & & 33.41 & ($-$0.28) & & 23.99 & ($+$0.03) & & 23.25 & ($+$0.60) \\
w/o $L_{\text{VAP}}$ & & 32.57 & ($-$1.12) & & 21.11 & ($-$2.85) & & 20.02 & ($-$2.63) \\
w/o $L_{\text{VAD}}$ & & 33.79 & ($+$0.10) & & 23.29 & ($-$0.67) & & 21.93 & ($-$0.72) \\
\hline
\end{tabular}
\end{table*}

\subsection{Setup}

We train three monolingual models (Japanese, English, Chinese) and one multilingual model (all three languages).
Details are as follows:
\begin{itemize}[topsep=2mm]
  \setlength{\itemsep}{0.2em}
  \setlength{\parskip}{0pt}
  \item \textbf{Model:} same architecture as Figure~\ref{fig:vap}; 1 transformer layer for each speaker and 3 cross-attention transformer layers; model dimension 256; 4 attention heads.
  Note that no language information (e.g., language ID) is explicitly provided to the model.
  \item \textbf{Dataset:} the corpus in Section~\ref{sec:dataset}; for each language, dialogues are randomly split into train (80\%), validation (10\%), and test (10\%).
  The multilingual model is trained on the combined training sets of all three languages.
  \item \textbf{Training:} AdamW optimizer; learning rate $3.63\times10^{-4}$, batch size 8, max 25 epochs.
  \item \textbf{Metric:} frame-level (100~ms) F1 score, following \cite{inoue2025naacl}, with a decision threshold of 0.5 on predicted probabilities.
\end{itemize}

\subsection{Cross-Lingual Performance}

Table~\ref{tab:f1_scores} presents F1 scores for monolingual and multilingual models.
As expected, monolingual models perform best on their own language (matched) but degrade substantially in zero-shot transfer.
This mirrors the cross-linguistic differences observed in Section~\ref{sec:dataset}.
For instance, a model trained on Japanese--where many backchannels occur during the speaker’s utterance--struggles on Chinese, where backchannels more often follow utterance completion, and vice versa.
English tends to fall between Japanese and Chinese both in frequency and silence timing, and accordingly shows intermediate cross-lingual transfer.
Nevertheless, the persistent degradation indicates that English does not simply subsume the other two languages.

By contrast, the multilingual model achieves performance comparable to or better than the matched monolingual models in all three languages.
This suggests that the model effectively learns universal cues while adapting to language-specific patterns based on the input.
The finding aligns with prior work on multilingual turn-taking prediction~\cite{inoue2024coling}.

We also evaluated two-language models in zero-shot settings (Table~\ref{tab:two_lang}).
These models, trained on pairs of languages, performed poorly when tested on the unseen third language, underperforming both the matched monolingual models and the three-language multilingual model.
This suggests that backchannel behaviors differ substantially across all three languages, making it difficult to learn universal and emergent prediction capabilities for backchannel behaviors.

Figure~\ref{fig:case_study} illustrates the behavior of the multilingual model across languages.
In (a) Japanese, the prediction probability rises just before true backchannel intervals, consistent with the relatively high F1.
In (b) English and (c) Chinese, while the peaks are less sharp, the model still captures backchannel timings reasonably well.
On the other hand, Figure~\ref{fig:case_study_jp} shows that the Japanese monolingual model struggles more with English and Chinese test data, producing many false positives and failing to capture backchannel timings.
This further highlights the limitations of monolingual models in cross-lingual settings.

\begin{figure*}[t]
\vspace{10pt}
\includegraphics[width=\linewidth]{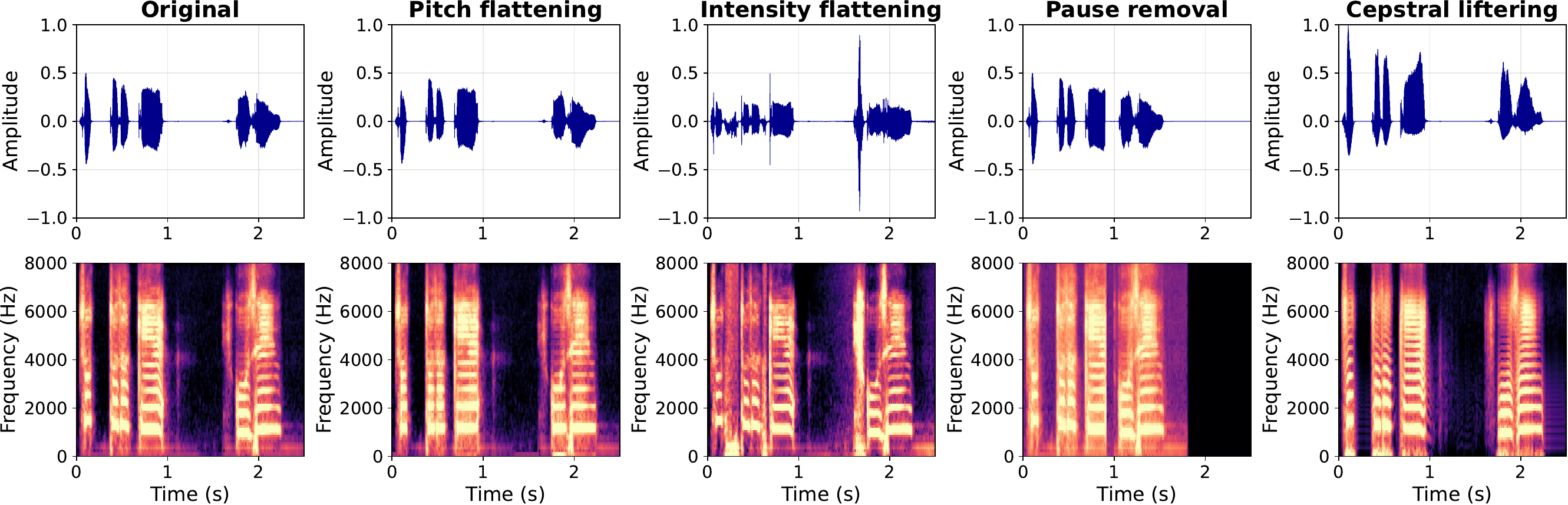}
\caption{Example of perturbation processing}
\label{fig:perturbation}
\end{figure*}

\subsection{Ablation Study}

We also conducted an ablation study to assess the contributions of auxiliary tasks (VAD, VAP, backchannel detection) to the main backchannel prediction task.
Table~\ref{tab:ablation_mono} shows the results for monolingual models.
Overall, we did not see any large degradation when removing auxiliary tasks, suggesting that monolingual models can learn backchannel prediction reasonably well on their own.
Rather, in some cases, removing auxiliary tasks slightly improved performance (e.g., removing $L_{\text{BD}}$ in Chinese and $L_{\text{VAD}}$ in Japanese).
This may be because the monolingual models can already capture language-specific cues effectively, and auxiliary tasks may introduce noise or conflicting signals.

On the other hand, the case of the multilingual model is different and showed consistent trends across languages.
Table~\ref{tab:ablation_multi} shows the results for the multilingual model.
Removing the VAP loss ($L_{\text{VAP}}$) causes the largest performance drop across all languages ($-$0.77 to $-$3.59 points), indicating that learning turn-taking dynamics also contributes to backchannel timing prediction.
Removing the backchannel detection loss ($L_{\text{BD}}$) also degrades performance, but to a lesser extent ($-$0.08 to $-$1.88 points), suggesting that explicit backchannel identification also aids prediction.
Interestingly, removing the VAD loss ($L_{\text{VAD}}$) slightly improves performance in all languages ($+$0.19 to $+$0.99 points), possibly because VAD may introduce noise when speech activity is not strongly correlated with backchannel timing prediction.
These findings indicate that auxiliary tasks play a more critical role in the multilingual setting, helping the model learn shared representations that generalize across languages.

\begin{table*}[t]
\centering
\caption{Perturbation analysis for (matched) \textbf{monolingual} models (F1 score [\%] and drop)}
\label{tab:perturb_mono}
\begin{tabular}{lcrrcrrcrr}
\hline
\multicolumn{1}{c}{Perturbation} & & \multicolumn{2}{c}{Japanese} & & \multicolumn{2}{c}{English} & & \multicolumn{2}{c}{Chinese} \\
\hline
None & & 33.27 & & & 22.85 & & & 21.37 & \\
Pitch flattening & & 31.84 & (\phantom{0}$-$1.43) & & 19.19 & (\phantom{0}$-$3.36) & & 20.46 & (\phantom{0}$-$0.91) \\
Intensity flattening & & 30.16 & (\phantom{0}$-$3.11) & & 19.99 & (\phantom{0}$-$2.86) & & 19.81 & (\phantom{0}$-$1.56) \\
Pause removal & & 30.32 & (\phantom{0}$-$2.95) & & 16.46 & (\phantom{0}$-$6.39) & & 15.32 & (\phantom{0}$-$6.05) \\
Cepstral liftering & & 17.02 & ($-$16.25) & & 9.58 & ($-$13.27) & & 5.45 & ($-$15.92) \\
\hline
\end{tabular}
\vspace{15pt}
\centering
\caption{Perturbation analysis for \textbf{multilingual} model (F1 score [\%] and drop)}
\label{tab:perturb_multi}
\begin{tabular}{lcrrcrrcrr}
\hline
\multicolumn{1}{c}{Perturbation} & & \multicolumn{2}{c}{Japanese} & & \multicolumn{2}{c}{English} & & \multicolumn{2}{c}{Chinese} \\
\hline
None & & 33.69 & & & 23.96 & & & 22.65 & \\
Pitch flattening & & 30.36 & (\phantom{0}$-$3.33) & & 19.53 & (\phantom{0}$-$4.43) & & 21.57 & (\phantom{0}$-$1.08) \\
Intensity flattening & & 27.38 & (\phantom{0}$-$6.31) & & 20.15 & (\phantom{0}$-$3.81) & & 19.85 & (\phantom{0}$-$2.80) \\
Pause removal & & 30.77 & (\phantom{0}$-$2.92) & & 17.57 & (\phantom{0}$-$6.39) & & 16.13 & (\phantom{0}$-$6.52) \\
Cepstral liftering & & 8.81 & ($-$24.88) & & 8.28 & ($-$15.68) & & 10.94 & ($-$11.71) \\
\hline
\end{tabular}
\end{table*}

\subsection{Perturbation Analysis}

To identify which input aspects the models exploit, we perform perturbation analyses by applying controlled manipulations to test audio and measuring performance changes.
As depicted in Figure~\ref{fig:perturbation}, we consider four perturbations:
\begin{itemize}[topsep=2mm]
    \setlength{\itemsep}{0.2em}
    \setlength{\parskip}{0pt}
    \item \textbf{Pitch flattening:} remove pitch variation to test reliance on F0 dynamics.
    \item \textbf{Intensity flattening:} flatten amplitude dynamics to test reliance on energy contours.
    \item \textbf{Pause removal:} remove up to 0.5~s of post-utterance silence to test reliance on silent gaps.
    \item \textbf{Cepstral liftering:} retain only low-order cepstral components to suppress phonetic content and test reliance on linguistic information.
\end{itemize}

Table~\ref{tab:perturb_mono} shows results for monolingual models.
The magnitude of degradation differs by language, indicating different feature usage.
The Japanese model is most affected by cepstral liftering ($-$16.25 points), suggesting strong reliance on linguistic information.
The English and Chinese models show large drops for both pause removal ($-$6.39 / $-$6.05) and cepstral liftering ($-$13.27 / $-$15.92), indicating sensitivity to both silence and linguistic cues.
The Chinese model is also relatively robust against pitch and intensity flattening ($-$0.91 / $-$1.56), suggesting less dependence on prosodic variation.

Results for the multilingual model (Table~\ref{tab:perturb_multi}) broadly follow similar trends, but with notable differences for Japanese and English: the impact of cepstral liftering increases to $-$24.88 and $-$15.68, respectively.
Note that the Chinese case shows a smaller drop ($-$11.71) compared to the monolingual one.
Pitch and intensity flattening also cause larger drops in the three languages, compared to the monolingual case, indicating increased reliance on prosodic cues.
This suggests that, when trained jointly, the model acquires a more language-aware strategy that emphasizes linguistic and prosodic information for all languages, while maintaining a similar level of sensitivity to silence cues.

\begin{table*}[t]
\centering
\caption{Context length analysis for (matched) \textbf{monolingual} models (F1 score [\%] and drop against 20 sec.)}
\label{tab:context_length_mono}
\begin{tabular}{ccrrcrrcrr}
  \hline
  \multicolumn{1}{c}{Context len. [sec.]} & & \multicolumn{2}{c}{Japanese} & & \multicolumn{2}{c}{English} & & \multicolumn{2}{c}{Chinese} \\
  \hline
  20 & & 33.27 & & & 22.85 & & & 21.37 & \\
  10 & & 33.75 & ($+$0.48) & & 22.39 & ($-$0.46) & & 22.11 & ($+$0.74) \\
  5 & & 33.18 & ($-$0.09) & & 22.49 & ($-$0.36) & & 21.63 & ($+$0.26) \\
  3 & & 33.46 & ($+$0.19) & & 21.60 & ($-$1.25) & & 19.52 & ($-$1.85) \\
  1 & & 32.72 & ($-$0.55) & & 18.52 & ($-$4.33) & & 11.90 & ($-$9.47) \\
  \hline
\end{tabular}

\vspace{15pt}

\centering
\caption{Context length analysis for \textbf{multilingual} model (F1 score [\%] and drop against 20 sec.)}
\label{tab:context_length_multi}
\begin{tabular}{ccrrcrrcrr}
  \hline
  \multicolumn{1}{c}{Context len. [sec.]} & & \multicolumn{2}{c}{Japanese} & & \multicolumn{2}{c}{English} & & \multicolumn{2}{c}{Chinese} \\
  \hline
  20 & & 33.69 & & & 23.96 & & & 22.65 & \\
  10 & & 33.63 & ($-$0.06) & & 24.25 & ($+$0.29) & & 22.31 & ($-$0.34) \\
  5 & & 33.48 & ($-$0.21) & & 24.20 & ($+$0.24) & & 22.79 & ($+$0.14) \\
  3 & & 33.20 & ($-$0.49) & & 23.00 & ($-$0.96) & & 19.96 & ($-$2.69) \\
  1 & & 32.15 & ($-$1.54) & & 20.58 & ($-$3.38) & & 15.69 & ($-$6.96) \\
  \hline
\end{tabular}
\end{table*}

\subsection{Context Length Dependency}

We further analyze how varying the input context length affects performance.
In the default setting, the model processes 20~s of past audio for both speaker and listener as the input context.
In this experiment, we again trained and used the multilingual model, but varied the input context length for the Transformer layers ranging from 1~s to 20~s during both training and inference.
Note that since the CPC encoder consists of CNN and GRU layers, it always processes the full 20~s input.
Table~\ref{tab:context_length_mono} and Table~\ref{tab:context_length_multi} present the results for monolingual and multilingual models, respectively.
Reducing the context length generally degrades performance, but the extent varies by language.
Japanese is relatively robust, with only a small drop ($-$0.55 and $-$1.54 points) even at 1~s context, suggesting that short-term cues suffice for backchannel prediction.
By contrast, English and Chinese show larger drops at 1~s context, indicating greater reliance on longer-term context.
Especially for Chinese, performance degrades sharply when context is reduced below around 3~s, suggesting that longer context is crucial for capturing relevant cues.
These differences may reflect language-specific conversational dynamics, such as the timing and distribution of backchannels.

\section{System Integration for Real-time Processing}

\begin{figure}[t]
\includegraphics[width=\linewidth]{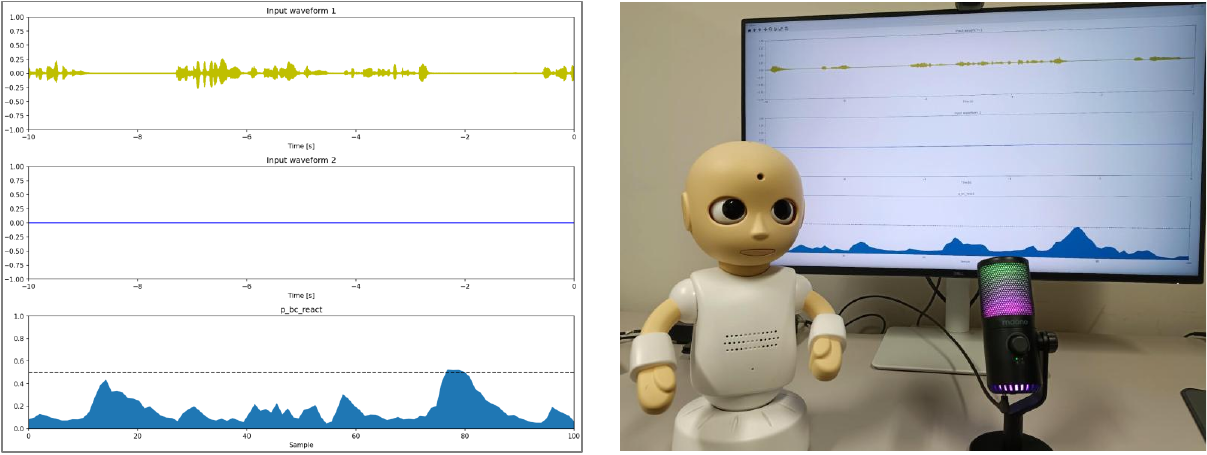}
\caption{Real-time processing software integrated with a dialogue robot}
\label{fig:maai}
\end{figure}

Finally, we integrated the trained backchannel prediction models into a real-time spoken dialogue system.
We implemented and released an open-source Python package, MaAI\footnote{\url{https://github.com/MaAI-Kyoto/MaAI}}, which supports real-time execution of VAP-based models (e.g., turn-taking, backchannel, and nodding prediction).
This package modularizes audio input/output (microphone, network, etc.), VAP processing, and visualization of VAP results, enabling straightforward integration into existing spoken dialogue systems and robots.
The trained backchannel prediction models are already integrated into the software; an example of its operation is shown in Figure~\ref{fig:maai}.
Thanks to the models' relatively small parameter counts and an efficient caching architecture, the 10~Hz backchannel predictor runs in real time on CPU only (Intel Core Ultra 9 285K).

\section{Conclusion}

We presented a multilingual backchannel prediction model for Japanese, English, and Chinese, enabling a cross-linguistic analysis of backchannel timing.
Trained on a large-scale multilingual conversational corpus, the proposed Transformer-based model achieved comparable or superior performance to monolingual models across all three languages.
These results demonstrate that the model successfully captures both language-universal cues and language-specific timing patterns.

Perturbation analyses revealed that the input cues underlying backchannel prediction differ across languages, highlighting distinct conversational strategies: Japanese listeners rely more on linguistic and short-term cues, while English and Chinese listeners are more sensitive to silence duration and prosodic variation.
The multilingual model effectively integrates these tendencies, suggesting that cross-lingual training encourages the emergence of shared yet adaptable representations of conversational feedback behavior.

Future work will focus on refining the annotation quality and expanding the scope of analysis.
Although utterance segmentation was performed manually, backchannel identification relied on ASR and surface-form matching; incorporating human-verified annotations would enable more precise modeling of backchannel types and functions.
We also plan to perform deeper interpretability analyses to elucidate the internal mechanisms by which the model captures language-universal backchannel cues.
Finally, by integrating the predictor into real-time spoken dialogue systems and evaluating it through human–machine interaction studies, we aim to quantify its impact on perceived naturalness, engagement, and conversational flow.

\section*{Acknowledgments}

This work was supported by JST PRESTO (JPMJPR24I4), JST Moonshot R\&D (JPMJPS2011), and JSPS KAKENHI (JP23K16901).

\bibliography{acl_latex}

\end{document}